\documentclass[conference]{IEEEtran}
\IEEEoverridecommandlockouts
\usepackage{cite}
\usepackage{amsmath,amssymb,amsfonts}
\usepackage{algorithmic}
\usepackage{graphicx}
\usepackage{textcomp}
\usepackage{xcolor}
\newsavebox{\measurebox}

\usepackage[caption=false]{subfig}
\usepackage{svg}
\usepackage{url}
\usepackage{multicol}
\usepackage{multirow}
\usepackage{color}
\usepackage{hhline}
\usepackage{pifont}
\usepackage{comment}
\usepackage{booktabs}
\usepackage{cleveref}
\usepackage{tablefootnote}
\usepackage{flushend}
\usepackage{romannum}
\usepackage[flushleft]{threeparttable}
\usepackage{colortbl}
\usepackage[ruled,vlined]{algorithm2e}
\definecolor{mynicegreen}{RGB}{11,102,35}
\definecolor{myniceblue}{rgb}{154,201,219}
\usepackage[group-separator={,}]{siunitx}

\newcommand{\model}{{\textsc{CLAUDIA}}}
\newcommand{\modelFullName}{{\textbf{C}ontrastive \textbf{L}earning with \textbf{A}uxiliary \textbf{U}ser \textbf{D}etection for \textbf{I}dentifying \textbf{A}ctivities}}

\DeclareMathOperator*{\argmin}{\arg\min}
\newcommand{\etal}{{\textit{et al.}}}

 \newcommand{\squishlist}{
	\begin{list}{$\bullet$}
		{ \setlength{\itemsep}{0pt}
			\setlength{\parsep}{3pt}
			\setlength{\topsep}{3pt}
			\setlength{\partopsep}{0pt}
			\setlength{\leftmargin}{1.5em}
			\setlength{\labelwidth}{1em}
			\setlength{\labelsep}{0.5em} } }
	
	\newcommand{\squishlisttwo}{
		\begin{list}{$\bullet$}
			{ \setlength{\itemsep}{0pt}
				\setlength{\parsep}{0pt}
				\setlength{\topsep}{0pt}
				\setlength{\partopsep}{0pt}
				\setlength{\leftmargin}{2em}
				\setlength{\labelwidth}{1.5em}
				\setlength{\labelsep}{0.5em} } }
		
		\newcommand{\squishend}{
	\end{list}}
\def\BibTeX{{\rm B\kern-.05em{\sc i\kern-.025em b}\kern-.08em
    T\kern-.1667em\lower.7ex\hbox{E}\kern-.125emX}}
    
\begin{document}

\title{Contrastive Learning with Auxiliary User Detection for Identifying Activities
\thanks{DARPA grant HR00111780032-WASH-FP-031 sponsored the research.}
}


\author{\IEEEauthorblockN{%
Wen Ge\textsuperscript{\textsection},
Guanyi Mou\textsuperscript{\textsection},
Emmanuel O. Agu,
Kyumin Lee}
\IEEEauthorblockA{\textit{Computer Science Department}, 
\textit{Worcester Polytechnic Institute}
Worcester, MA, USA \\
{\{wge, gmou, emmanuel, kmlee\}}@wpi.edu}
}



\maketitle
\begingroup\renewcommand\thefootnote{\textsection}
\footnotetext{Equal contribution}
\endgroup

\begin{abstract}
Human Activity Recognition (HAR) is an important task in ubiquitous computing, with impactful real-world applications. While recent state-of-the-art HAR research has demonstrated impressive performance, some key aspects remain under-explored. First, we believe that for optimal performance, HAR models should be both Context-Aware (CA) and personalized. However, prior work has predominantly focused on being Context-Aware (CA), largely ignoring being User-Aware (UA). We argue that learning user-specific differences in performing various activities is as critical as considering user context while performing HAR tasks. Secondly, we believe that the predictions of HAR models should be unified, reliably recognizing the same activity even when performed by different users.
As such, the representations utilized by CA and UA models should explicitly place different users performing the same activity closer together. Moreover, identifying the user performing an activity is useful in applications such as thwarting cheating by having another person perform medically-prescribed activities.   

To bridge this gap, we introduce \textbf{\modelFullName} (\model), a novel framework designed to address these issues. Specifically, we expand the contextual scope of the CA-HAR task by integrating User Identification (UI) within the CA-HAR framework, jointly predicting both CA-HAR and UI in a new task called User and Context-Aware HAR (UCA-HAR). This approach enriches personalized and contextual understanding by jointly learning user-invariant and user-specific patterns. Inspired by state-of-the-art designs in the visual domain, we introduce a supervised contrastive loss objective on instance-instance pairs to enhance model efficacy and improve learned feature quality. Through theoretical exposition, empirical analysis of real-world datasets, and rigorous experimentation, we demonstrate the significance of each component of {\model} and discuss its relationship with existing methodologies. Evaluation across three real-world CA-HAR datasets reveals substantial performance enhancements, with average improvements ranging from 11.7\% to 14.2\% in Matthew's Correlation Coefficient (MCC) and 5.4\% to 7.3\% in Macro F1 score. To encourage further research, we share code with additional supplement material in repository {\url{https://github.com/GMouYes/CLAUDIA}}.

\end{abstract}


\section{Introduction}

Human Activity Recognition (HAR), which identifies activities such as walking or running from sensor data, plays a crucial role in health and medical applications \cite{zhou2020deep, wan2020deep, chen2018evaluating, lindqvist2011undistracted}. HAR is challenging due to co-occurring activities (e.g., talking while walking), making it a multi-label classification task that requires modeling these relationships.

Human activities are highly \textit{contextualized}; people perform different activities in different contexts/situations. Consequently, recent studies have increasingly focused on Context-Aware Human Activity Recognition (CA-HAR) \cite{lara2012survey, ronao2016human, hassan2018robust} as a promising approach. The fundamental premise of CA-HAR lies in acknowledging and capturing the significant impact of context on sensor signals by integrating context classification as an auxiliary task \cite{vaizman2018context, ge2023heterogeneous, cao2018gchar}. However, another crucial perspective is often overlooked: human activities can also be highly \textit{personalized}. The variety and heterogeneity in human behavior inevitably lead to variances in action-performing patterns \cite{lane2011enabling,bull2020world}, even for the same activity. HAR frameworks focusing solely on external contextual settings may easily overlook the innate differences across users performing the activities, potentially leading to erroneous results. Inspired by this observation, our work aligns with the field of CA-HAR research while also extending beyond its scope by devising a novel framework that accounts for both external contexts and internal user identities. Specifically, we incorporate User Identification (UI) as an auxiliary subtask, effectively extending traditional context-aware (CA) HAR design into user-context-aware (UCA) HAR. Identifying the user performing an activity is useful in applications such as thwarting  cheating by having another person perform medically-prescribed activities.  

Adding UI as a subtask enables the model to acknowledge user identity but does not necessarily guarantee overcoming inter-user differences. To address the challenge of user-specific variability, we introduce {\modelFullName} (\model) with a supervised contrastive loss that regularizes (different-user, same-activity) representations, ensuring closer vector similarity and capturing both user-specific and user-invariant information effectively.

In summary, we propose a novel framework \textbf{\modelFullName} (\model) that incorporates User Identification (UI) as a subtask alongside the primary CA-HAR task, explicitly regularizing the modeling process using an additional loss objective. This User-Context-Aware HAR (UCA-HAR) framework achieves a balance between capturing user-invariant patterns and user-specific attributes in a single model. Drawing inspiration from inter-instance relationship modeling in computer vision and natural language processing domains, we introduce a novel supervised contrastive loss on instance-instance pairs. By leveraging inter-instance relationships, the {\model} framework bridges the gaps observed in current CA-HAR approaches, resulting in refined instance representations.
This research offers the following contributions:
\begin{itemize}
\item We extend the context definition in CA-HAR by incorporating a UI subtask to capture both user-invariant and user-specific patterns, improving HAR performance.
\item We propose the {\model} framework, which integrates multi-modality sensor data and a novel supervised contrastive loss to enhance HAR.
\item Extensive experiments demonstrate that {\model} outperforms state-of-the-art baselines across several datasets, showcasing its adaptability and effectiveness.
\end{itemize}

\section{Related Work}
\label{sec:relatedwork}
This section provides an overview of related work, including prior CA-HAR model designs, recent advancements in the UI task, and research related to contrastive learning. 

\subsection{CA-HAR Models}

Prior CA-HAR work primarily focused on two aspects: 1) Directly modeling instance-label relationships and 2) Enhancing (1) by auxiliary modeling of label-label relationships.

\subsubsection{Directly Modeling Instance-Label Relationships}

Early CA-HAR models used handcrafted sensor features with traditional or deep learning methods\cite{gao2019human, vaizman2018context}. However, manual feature extraction is labor-intensive and lacks generalizability. Recently, Deep Neural Networks (DNNs) have shown success by auto-extracting features, using architectures such as CNNs \cite{munzner2017cnn}, RNNs \cite{mekruksavanich2021lstm}, and Graph Neural Networks (GNNs)\cite{shi2020skeleton}. Though DNNs offer improved generalization, they often lack interpretability, which is addressed in recent works combining learned and handcrafted features \cite{ge2020cruft, ge2022qcruft}.

\subsubsection{Enhanced Auxiliary Modeling of Inter-Label Relationships}

To capture label relationships, some studies transformed the task into graph learning, such as HAR-GCNN~\cite{mondal2020new}, which uses graph CNNs, and HHGNN~\cite{ge2023heterogeneous}, encoding user, context, and activity labels as nodes and co-occurrences as hyperedges. These approaches enhanced HAR performance by capturing label dependencies.
%
%
Despite these advances, inter-instance relationships remain under-explored in CA-HAR research. Our work introduces supervised contrastive learning on instance pairs to address this, improving feature quality (Section~\ref{sec:framework}).

\subsection{User Identification and Applications}

User identification has gained importance with the rise of smartphones. Zou \etal\cite{zou2020deep} used CNN and LSTM to extract gait features for user identification, while Mekruksavanich and Jitpattanakul\cite{mekruksavanich2021biometric} developed an ensemble model combining activity and user classifiers. However, these methods struggle with real-world complexity. Our unified UCA-HAR model integrates UI with CA-HAR, showing superior performance across a wide and diverse range of activity scenarios, demonstrating its versatility and potential.

\subsection{Contrastive Loss for HAR}
We apply supervised contrastive learning to regularize inter-instance relationships and enhance feature quality. Unlike self-supervised contrastive methods relying on augmentation or external data\cite{tang2020exploring, wang2022sensor}, our approach utilizes labeled data to achieve improved performance~\cite{khosla2020supervised}.

\begin{figure*}
    \centering
    \vspace{-10pt}
    \includegraphics[width=.75\linewidth]{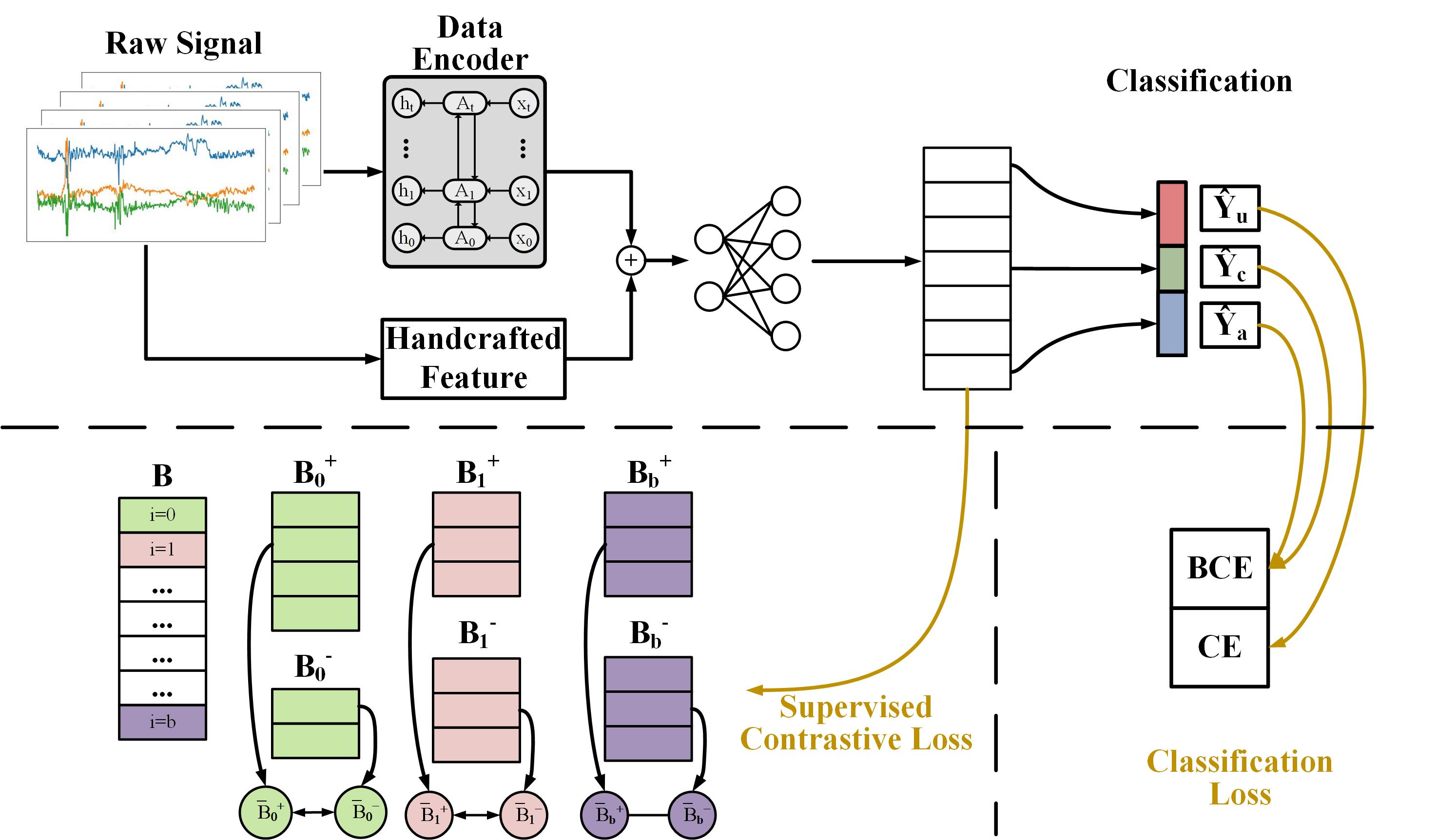}
    \caption{Framework of {\model}.The upper part illustrates the network design, comprising of data encoding and classification. The lower part depicts the objective function, consisting of two types of losses. Our technical innovation lies in two key aspects: 1) the addition of a UI sub-task, thereby broadening the scope of Context-Aware (CA) HAR to effectively include User-Aware (UA) HAR, and 2) a novel approach to improve feature quality by regularizing inter-instance relationships using supervised contrastive loss.}
    \label{fig:enter-label}
    \vspace{-10pt}
\end{figure*}

\section{Proposed {\model} Framework}
\label{sec:framework}

We describe our novel framework in the following subsections. We first illustrate the task formulation in Section~\ref{sec:task}. We then demonstrate a high-level picture of our design in Sec.~\ref{sec:overview}, followed by each sub-module. Lastly, we introduce the loss objectives in Sec.~\ref{sec:loss}, including a UI sub-task loss.

\subsection{Task Formulation}
\label{sec:task}
Given input raw data $X_{raw} \in R^{N*S*T}$, where \( N \) represents the number of samples, \( S \) denotes the number of sensors, and \( T \) indicates the snapshots per sample, we compute handcrafted features $f$, such that:
\begin{equation}
\label{eq:hc}
X_{hc} = f(X_{raw}), f: R^{S*T} \rightarrow R^D
\end{equation}
%
%
%
$X_{hc}$ provides a more compact representation, where \( D \) $\ll$ \( S \times T \). Traditional HAR methods $M_{\theta}$ map input to labels:
\begin{equation}
    \label{eq:original_task}
    M_{\theta}: X_{raw/hc} \rightarrow Y, 
\end{equation}
%
Our innovative performance of the UI sub-task, along with our design with supervised contrastive loss objective on inter-instance relationship modeling, are our unique contributions.

\subsection{Model Overview}
\label{sec:overview}
$\model$ $M$ contains three components (Fig.~\ref{fig:enter-label}): 

\smallskip\noindent\underline{1) \textit{Data encoder $M_{\theta_{t}}$}:} Extracts features from $(X_{raw}, X_{hc})$, producing $X_r$ (Eq.~\ref{eq:dataEncoder}).
\begin{equation}
    \label{eq:dataEncoder}
    X_r = M_{\theta_{t}}(X_{raw}) \oplus X_{hc}, X_r \in R^{N*d_t}
    \vspace{-5pt}
\end{equation}
where $\oplus$ is feature concatenation and $d_t$ is the new feature dimension. We used a two-layer LSTM for sequence learning.

\smallskip\noindent\underline{2) \textit{Label encoder}:} Encodes activities, user identities, and context settings into binary labels $Y$. 
While there are other methods beyond 0/1 label encoding \cite{ge2023heterogeneous,ge2024deep,kipf2016semi,bai2021hypergraph}, we test with the straightforward method based on its simplicity and note these methods are supplemental to our design.

\smallskip\noindent\underline{3) \textit{Alignment module} $M_{\theta_{cls}}$:} Aligns encoded data with label representations using separate linear projection followed by af 
  ctivation for each label type, resulting in predictions $\hat{Y}$, including activity $\hat{Y}_A$, phone placement $\hat{Y}_{PP}$, and UI $\hat{Y}_U$.

%


\subsection{Objective Function}
\label{sec:loss}
Our overall objective minimizes the total loss $L$, which is composed of the classification loss $L_{cls}$ and the contrastive loss $L_d$, weighted by $\alpha$:
%
%
\begin{equation}
    \label{eq:overallLoss}
    \begin{aligned}
        \hat{\theta} &= \argmin_{\theta} L(M_{\theta}, X_{raw}, Y) \\
        L &= L_{cls} + \alpha L_{d}
    \end{aligned}
\end{equation}
\noindent\textbf{Classification Loss}:
$L_{cls}$ includes activity $L_A$, phone placement $L_{PP}$, and UI $L_U$ tasks, weighted by $\gamma_1$ and $\gamma_2$:
%
\begin{equation}
    \label{eq:clsLoss}
    L_{cls} = L_A + \gamma_1 L_{PP} + \gamma_2 L_{U}
\end{equation}
As activities and general contexts may co-occur, we leverage Binary Cross Entropy (BCE) for $L_A, L_{PP}$ (Eq.~\ref{eq:bce}). In this case, class indices range from $0$ to $C-1$, where $C$ represents the total number of classes. $\hat{y}{n,c}$ and $y{n,c}$ denotes the predicted and ground truth label, respectively, while $\omega_{n,c}$ represents the class weight, which is calculated based on the inverse frequency ratio to address the class imbalance issue. On the other hand, as user identities are usually mutually exclusive, we adopted general Cross Entropy (CE) for $L_U$ (Eq.~\ref{eq:ce}.)
\begin{equation}
    \label{eq:bce}
    \resizebox{0.88\hsize}{!}{$
    L_{bce} = -\omega_{n,c}[y_{n,c} \cdot log \sigma(\hat{y}_{n,c}) + (1-y_{n,c})\cdot log(1-\sigma(\hat{y}_{n,c}))]
    $}
\end{equation}
\begin{equation}
    \label{eq:ce}
    L_{ce} = -\sum_{c=1}^Clog\frac{exp(\hat{y}_{n,c})}{\sum_{i=1}^Cexp(\hat{y}_{n,i})}y_{n,c}
\end{equation}
\noindent\textbf{Instance-Pair Supervised Contrastive Loss:}
%
We introduce our novel design for the sample-based supervised contrastive loss in the data encoder. Given any chosen anchor $x_a, y_a$ in any batched samples $B$ with $X_b, Y_b$ in a training step, we first separate the batch into positive pairs and negative pairs:
\begin{equation}
\label{eq:posNegPairs}
\begin{aligned}
    B &= \{(x,y) ~|~ x = X_b[i], y = Y_b[i], i=0,1,...\} \\
    B_a^{+} &= \{(x,y) ~|~ y \cdot y_a^T\ > 0, (x,y) \in B\} \\
    B_a^{-} &= \{(x,y) ~|~ y \cdot y_a^T == 0, (x,y) \in B\} \\
    B &= B_a^{+} \cup B_a^{-}, \emptyset = B_a^{+} \cap B_a^{-}
\end{aligned}
\end{equation}
Essentially, the positive pairs contain instances that share at least one label (i.e., non-zeros) with the anchor instance, while the negative pairs share no label with the anchor. Note that the size of both $B_a^{+}, B_a^{-}$ can be volatile even under a fixed batch size. The changing size directly impedes applying a contrastive loss similar to Contrastive Language-Image Pre-training (CLIP)~\cite{radford2021learning}. To accommodate the issue, we calculated the average of both positive and negative pairs:
%
\begin{equation}
    \label{eq:sampling}
    \begin{aligned}
        x_a^{+} &= \frac{1}{|B_a^{+}|} \sum \{x \text{~for~} (x,y) \in B_a^{+}\} \\
        x_a^{-} &= \frac{1}{|B_a^{-}|} \sum \{x \text{~for~} (x,y) \in B_a^{-}\}
    \end{aligned}
\end{equation}
When the negative pair set is empty, we utilize a zero vector to account for its average. Another option is to sample fixed numbers $k^+, k^-$ of the positive and negative sets. However, such a method would necessitate an undersampling and oversampling design, and would inevitably involve an extra hyperparameter search for  $k^+, k^-$. Taking these factors into consideration, we opted for a computationally more straightforward choice as mentioned. We leave the exploration of alternate sampling methods as future work for interested readers. 

Using the supervised contrastive loss function, we pull the anchor $x_a$ towards the positive sample $x_a^{+}$ and enforce the distances between anchor $x_a$ and the negative sample $x_a^{-}$ even further away. Such an optimization direction can be reached via an cross-entropy loss:
\begin{equation}
    \label{eq:scl}
    \begin{aligned}
        p_{x_a} &= [sim(x_a, x_a^{+}), sim(x_a, x_a^{-})] \\
        q_{x_a} &= [1, 0] \\
        L_d &= \frac{1}{|B|}\sum_{x \in B} CE(p_{x_a}, q_{x_a})
    \end{aligned}
\end{equation}
We re-emphasize that while prior work in the HAR domain mainly focused on classification loss (instance-label mapping regulation), we propose explicitly regularizing inter-instance relationships using a supervised contrastive loss.

\begin{table*}
    \centering
    \vspace{-5pt}
    \small
    \caption{Comparison of 3 benchmark CA-HAR datasets. The accelerometer, gyroscope, magnetometer, and gravity are tri-axial sensors. Sampling rate is once per minute or when its value changed more than a threshold value if no sampling rate was specified.}
    \vspace{-5pt}
    \scalebox{.85}{
    \begin{tabular}{l|c|c|c}
    \toprule  
    \textbf{Dataset}&   \textbf{\textit{WASH scripted}} &\textbf{\textit{WASH unscripted}}  &\textbf{\textit{Extrasensory}}\\
    \midrule
    \textbf{Category}&   \textit{Scripted} &\textit{Unscripted} &\textit{Unscripted} \\
    
    \textbf{Instances}         & \num[group-separator={,}]{294512}  &  \num[group-separator={,}]{7773479} &  \num[group-separator={,}]{6355350} \\
    
    \textbf{Features}          & \num[group-separator={,}]{144}  & \num[group-separator={,}]{139}  & \num[group-separator={,}]{170}  \\
    \textbf{Participants}      & \num[group-separator={,}]{107}  & \num[group-separator={,}]{108}  & \num[group-separator={,}]{60}  \\
    \textbf{Contexts}    & \num[group-separator={,}]{5}  &  \num[group-separator={,}]{5} &  \num[group-separator={,}]{4} \\
    \textbf{Activities}   & \num[group-separator={,}]{12}  &  \num[group-separator={,}]{12} &  \num[group-separator={,}]{12} \\\midrule
    \multirow{2}{*}{\textbf{Common Sensors}} &\multicolumn{3}{c}{\multirow{2}{*}{\shortstack{Accelerometer(40Hz), Gyroscope(40Hz), Magnetometer(40Hz),\\Location, Environment Measure, Phone State}}}\\
    \\\midrule
    \textbf{Unique Sensors} &   &  & Gravity(40Hz), Audio(46Hz)\\
    \midrule
    \textbf{Context Labels} &  \multicolumn{3}{c}{In Pocket, In Hand, In Bag}\\\midrule
    \textbf{Unique Contexts}& On Table & On Table & On Table-Face Down/Up\\\midrule
    \textbf{Common Activities} & \multicolumn{3}{c}{Lying Down, Sitting, Walking, Sleeping, Standing, Running, Stairs-Going Down/Up}\\\midrule
    \textbf{Unique Activities}  & \multicolumn{2}{c|}{Talking on Phone, Bathroom, Jogging, Typing} & Talking, Bath-Shower, Toilet, Exercising\\
    \bottomrule
    \end{tabular}}
    \label{tab:datasets}
    \vspace{-10pt}
\end{table*}

\section{Experiments}
\label{sec:experiment}
We introduce real-world CA-HAR datasets in Section~\ref{sec:dataset} and describe state-of-the-art baselines in Section~\ref{sec:baselines}. We then describe our experimental setup (Sec.~\ref{sec:exp_setup}), evaluation metrics (Sec.~\ref{sec:metrics}), and experimental results (Sec.~\ref{sec:results}).

\subsection{Datasets}
\label{sec:dataset}

{\model} was evaluated on three CA-HAR datasets, encompassing scripted and unscripted data collection study designs: \textit{WASH}\footnote{\url{https://tinyurl.com/darpaWash}} (Scripted and Unscripted) and \textit{Extrasensory}\cite{vaizman2017recognizing} (Unscripted) with detailed statistics in Table~\ref{tab:datasets}.
In scripted datasets, subjects perform activities as instructed, visiting specific pre-defined target labels in a constrained order. In contrast, unscripted datasets allow participants live their lives unconstrained, then annotate their data periodically.  Unlike prior studies focusing on either solely on scripted or unscripted data, we evaluated both to gain comprehensive insights. We predicted 12 activity labels common to all three datasets, ensuring a fair comparison with previous work~\cite{asim2020context,gong2022note,cruciani2020feature}.

\textit{Pre-processing:} Instances with conflicting labels (e.g., sleeping while running) were removed (13\% of \textit{WASH unscripted} and 34\% of \textit{Extrasensory} data). Sensor readings were segmented with a 3-second window and 1.5-second step.

\textit{Feature extraction and selection:} We extracted handcrafted features ($f$) identified as highly predictive in prior work~\cite{vaizman2018context,ge2020cruft,ge2022qcruft,ge2023heterogeneous,ge2024deep}, yielding 144, 139, and 170 features for the three datasets, respectively. Sensor signals were resampled into fixed-length (50 samples) sequences using the Fourier method~\cite{cooley1965algorithm}, resulting in $x_{raw} \in \mathcal{N}^{12\times50}$. Features were normalized using training set statistics (i.e., $x = (f-\mu)/s$) with missing values filled with zeros.
%
%
%
%
%


\subsection{Baseline HAR Models}
\label{sec:baselines}
We compared {\model} to a diverse set of state-of-the-art CA-HAR models, covering graph-based (GCN~\cite{kipf2016semi}, HGCN~\cite{bai2021hypergraph}, HHGNN~\cite{ge2023heterogeneous}), non-graph-based (ExtraMLP~\cite{vaizman2018context}, LightGBM~\cite{ke2017lightgbm}, CRUFT~\cite{ge2020cruft}, GaitAuth~\cite{zou2020deep}, GaitIden~\cite{zou2020deep}), deep learning (ExtraMLP, CRUFT, GaitAuth, GaitIden), and machine learning (LightGBM) techniques. Due to space limits, we report the detailed descriptions of each model in our supplement materials (available in our repository).

\begin{figure*}
    \centering
    \vspace{-10pt}
    \includegraphics[width=.95\linewidth]{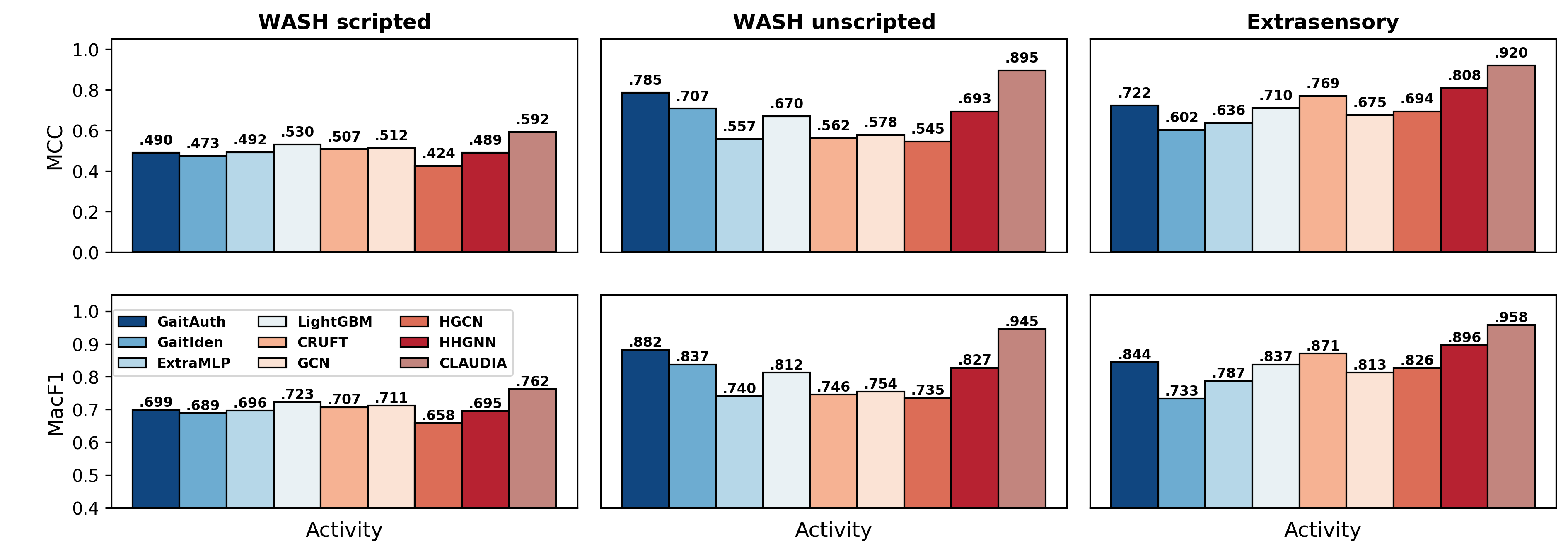}
    \vspace{-10pt}
    \caption{Average activity recognition performance of all models across all datasets. }
    \label{fig:activityPerfomance}
    \vspace{-10pt}
\end{figure*}

\subsection{Experimental Setup}
\label{sec:exp_setup}

We split data with training (60\%), validation (20\%), and testing (20\%) for all users. Validation set was used to determine optimal model hyperparameters via grid search. For {\model}, we used the RAdam~\cite{liu2019variance} optimizer, and set batch size = 1024.  
We share hyperparameters in our repository.

\begin{table*}
    \centering
    \small
    \caption{Model performance on the \textit{WASH unscripted} dataset, each cell contains MCC/Macro-F1 scores.}
    \vspace{-5pt}
    \scalebox{.84}{
    \begin{tabular}{c||c|c|c|c|c|c|c|c|c|r}
    \toprule
    Category    &   \multicolumn{2}{c|}{UI/UA}   & \multicolumn{3}{c|}{non-GNN}   & \multicolumn{3}{c|}{GNN}& \multicolumn{2}{c}{Our Model}\\\midrule
    \textbf{Model}& \textbf{GaitAuth} & \textbf{GaitIden} & \textbf{ExtraMLP}& \textbf{LightGBM} & \textbf{CRUFT} & \textbf{GCN} & \textbf{HGCN} & \textbf{HHGNN} & \textbf{\model} & \textbf{Improv.(\%)}  \\
    \midrule
    Lying Down              & .842/.916 & .810/.899 & .831/.912 & .783/.887 & .820/.907 & .848/.921 & .729/.859 & .893/.945  & .975/.987 & 9.2/4.4\\
    Sitting                 & .886/.942 & .820/.908 & .741/.867 & .735/.864 & .758/.876 & .743/.868 & .684/.838 & .816/.907  & .961/.980 & 8.5/4.1\\
    Walking                 & .808/.899 & .698/.838 & .523/.731 & .536/.741 & .510/.721 & .519/.732 & .498/.716 & .623/.794  & .891/.945 & 10.3/5.1\\
    Sleeping                & .938/.969 & .916/.958 & .921/.960 & .922/.961 & .915/.957 & .935/.967 & .927/.963 & .954/.977  & .989/.994 & 3.6/1.8\\
    Talking On Phone        & .767/.872 & .620/.779 & .436/.656 & .545/.730 & .417/.645 & .465/.675 & .489/.695 & .657/.806  & .875/.935 & 14.1/7.2\\
    Bathroom                & .592/.761 & .552/.740 & .427/.659 & .499/.704 & .425/.660 & .406/.641 & .481/.694 & .615/.780  & .817/.902 & 32.7/15.6\\
    Standing                & .751/.864 & .636/.795 & .463/.686 & .486/.703 & .457/.681 & .489/.705 & .450/.677 & .605/.777  & .889/.943 & 18.3/9.1\\
    Jogging                 & .836/.912 & .749/.862 & .520/.712 & .964/.982 & .551/.737 & .599/.765 & .409/.641 & .712/.837  & .922/.960 & -4.3/-2.2\\
    Running                 & .712/.839 & .677/.821 & .420/.647 & .946/.973 & .479/.692 & .489/.695 & .344/.604 & .604/.769  & .881/.938 & -6.9/-3.6\\
    Stairs-Going Down       & .700/.833 & .806/.780 & .376/.626 & .488/.697 & .382/.636 & .374/.624 & .516/.721 & .529/.726  & .806/.897 & 15.1/7.7\\
    Stairs-Going Up         & .683/.822 & .571/.752 & .388/.634 & .469/.686 & .397/.645 & .399/.640 & .545/.740 & .537/.731  & .791/.888 & 15.9/8.0\\
    Typing                  & .900/.948 & .832/.912 & .636/.794 & .666/.814 & .636/.793 & .672/.817 & .467/.670 & .770/.876  & .949/.974 & 5.5/2.7\\

    \bottomrule
    \end{tabular}}
    \vspace{-5pt}
    \label{tab:main_exp_unscripted}
\end{table*}

\subsection{Evaluation Metrics}
\label{sec:metrics}
We used Matthews Correlation Coefficient (MCC) (Eq.\ref{eq:mcc}) \cite{chicco2020advantages} and Macro-F1 scores (Eq.~\ref{eq:f1}) for imbalanced CA-HAR datasets. MCC reflects model performance across all labels, accounting for data imbalance. The Macro-F1 score, calculated as the average F1 across all labels, combines precision ($Precision=\frac{TP}{TP+FP}$) and recall ($Recall=\frac{TP}{TP+FN}$), providing a comprehensive evaluation.
%
%
\begin{equation}
    \resizebox{0.8\hsize}{!}{$MCC = \frac{TP \times TN - FP \times FN}{\sqrt{(TP+FP)(TP+FN)(TN+FP)(TN+FN)}}$}
    \label{eq:mcc}
\end{equation}
\begin{equation}
\resizebox{0.85\hsize}{!}{$F1 = 2*({Precision_{c_i}*Recall_{c_i}})/({Precision_{c_i}+Recall_{c_i}})$}
\label{eq:f1}
\end{equation}

\subsection{Experiment Results}
\label{sec:results}

\subsubsection{Main HAR Results}
\label{sec:exp_har}
We present comprehensive results of evaluation on HAR performance in Fig~\ref{fig:activityPerfomance}.

\smallskip\noindent\underline{Overall Performance:}
{\model} outperforms all baseline models across all datasets. Specifically, {\model} outperforms the best baselines by an average of 11.7\%, 14.2\%, and 13.8\% in MCC on the \textit{WASH scripted}, \textit{WASH unscripted}, and \textit{Extrasensory} dataset, respectively; When measured with Macro-F1 scores, it also improved 5.4\%, 7.3\%, and 7.0\% against the best baselines.

\smallskip\noindent\underline{Noteworthy Models and Labels:} 

We present the detailed results for \textit{WASH unscripted} in Table~\ref{tab:main_exp_unscripted}, with other dataset results available in our repository and supplementary materials (which follow similar trends). Notably, {\model} showed significant improvement in recognizing activities like \textit{Walking} and \textit{Bathroom/Toilet}, which involve periodic patterns and varied user movements, often with weaker labeling due to privacy concerns—aligning well with {\model}'s strengths. Additionally, our model was also impressive on \textit{Stairs-Going Up/Down}, which are challenging to distinguish~\cite{ravi2005activity}.

Although LightGBM had an edge on \textit{Jogging} and \textit{Running} (inherently similar in nature and could confound each other), {\model} remains the best among unified models.

\smallskip\noindent\underline{Comparative Analysis across Datasets:}
Generally speaking, most models exhibited higher performance on the \textit{WASH Unscripted} and \textit{Extrasensory} datasets compared to the \textit{WASH Scripted} dataset. This could be attributed to the \textit{WASH Scripted} having the least number of instances and a lower instance-to-user ratio, as reported in Table~\ref{tab:datasets}.

\begin{table}[ht]
    \centering
    \small
    \vspace{-5pt}
    \caption{Ablation study compares our \model model with its variants for the activity recognition task.}
    \vspace{-5pt}
    \scalebox{.9}{
    \begin{tabular}{r||c|c|c}
    \toprule
    {Model}   & {\textit{WASH scripted}}  & {\textit{WASH unscripted}} & {\textit{Extrasensory}}\\
    \midrule
    {\model} &    .592/.762 & .895/.945 & .920/.958\\
    w/o UI & .487/.696 & .871/.932 & .909/.952\\
    w/o CL & .574/.751 & .873/.933 & .899/.947\\
    w/o TS & .478/.688 & .764/.870 & .843/.915\\
    \bottomrule
    \end{tabular}}
    \label{tab:ablation}
    \vspace{-5pt}
\end{table}

\subsubsection{Ablation on {\model}}
\label{sec:ablate}

We conducted an ablation study on {\model} to assess the contributions of each modules. Table~\ref{tab:ablation} presents the corresponding MCC and Macro-F1 scores for the full model and its variants:
%
1) w/o UI: {\model} without the user identification module ($\gamma_2$ = 0);
%
%
2) w/o Contrastive Loss (CL): {\model} without instance-level contrastive regularization ($\alpha$ = 0);
3) w/o Time-Series (TS) encoding: Omits the sequence learning model, using only handcrafted features.
The results demonstrate that the performance of {\model} deteriorates when any of these key components is removed, underscoring the contribution of each module to its performance. Notably, the time-series module has the most significant impact especially on unscripted datasets. Integrating them enables {\model} to achieve robust results across both scripted and unscripted datasets.


%

\section{Analysis}
\label{sec:analysis}

\subsection{Context Identification}
\label{sec:context_cls}

{\model} achieves MCC/Macro-F1 scores 0.813/0.900, 0.946/0.973, and 0.974/0.987 on the \textit{WASH scripted}, \textit{WASH unscripted}, and \textit{Extrasensory} datasets respectively. Detailed confusion matrices are provided in the supplementary material.

\subsection{User Identification}
\label{sec:user_cls}

%
%
%
{\model} demonstrated near-perfect UI performance, achieving MCC and Macro-F1 scores of 0.995, 0.999, and 1.000 on the three datasets. It significantly outperformed \textit{GaitIden} (with .959/.959, .962/.982, and .965/.963 MCC/Macro-F1 scores), especially on \textit{WASH scripted}, indicating {\model}'s effectiveness even when user distinctions were less clear.

\section{Conclusion and Future Work}
\label{sec:conclusion}
In this work, we improve Context-Aware Human Activity Recognition (CA-HAR) task performance by 1) expanding the contextual scope to include user identification and in learning user-specific patterns, and 2) introducing a novel inter-instance relationship objective that supplements existing instance-label modeling targets by enforcing user-invariant activity embeddings.
We propose {\model}, a framework that leverages both handcrafted feature and raw time-series data, innovatively performs user identification as an auxiliary sub-task to the CA-HAR framework. The resulting framework, \model, significantly improves HAR performance across multiple real-world CA-HAR datasets with average enhancements ranging from 11.7\% to 14.2\% in Matthew’s Correlation Coefficient (MCC) and 5.4\% to 7.3\% in Macro F1 score. 
Furthermore, we provide extensive qualitative and quantitative analysis results. Since we only experimented with recurrent-based networks for time series modeling, we leave the exploration of other architectures, such as Transformers, for future work.

\bibliographystyle{IEEEtran}
\bibliography{refs}
\flushend
\end{document}